\documentclass[runningheads]{llncs}

\usepackage{accv}

\usepackage{accvabbrv}

\usepackage{graphicx}
\usepackage{booktabs}
\usepackage{amsmath}
\usepackage{amssymb}
\usepackage{amsfonts}
\usepackage{multirow}
\usepackage{placeins}    

\usepackage[accsupp]{axessibility}  

\usepackage{hyperref}

\usepackage{orcidlink}

\graphicspath{{./}{figures/}}

\begin{document}

\title{dKFD: Phase-Structured Evidence Allocation for Fixed-Budget Localized Event Understanding}

\titlerunning{dKFD: Phase-Structured Evidence Allocation}

\author{Aditya Bagri\inst{1} \and
Ashutosh Kumar\inst{2}\and
Chaitanya Lakhchaura\inst{1} \and
Avinash Anand\inst{3} \and
Zhengkui Wang\inst{3} \and
Rajiv Ratn Shah\inst{5}
}

\authorrunning{A. Bagri et al.}

\institute{Indraprastha Instituition of Information Technology, New Delhi, India 
\and
Rochester Institute of Technology, NY, USA
\and
SIT × Nvidia AI Center (SNAIC), Singapore
\and
NVIDIA, Singapore
\and
Indian Institute of Technology Kanpur, Kanpur, India
}

\maketitle

\begin{abstract}
Sparse video understanding often requires selecting a small set of visual evidence under a fixed frame budget. Most sparse selectors allocate this budget globally, allowing all frames to compete with one another. For temporally localized events, this can be a poor inductive bias: useful evidence is often distributed across pre-event context, the event itself, and post-event consequences. We study fixed-budget evidence allocation for localized event videos and show that globally competitive Top-$K$ selectors can preserve recognition and grounding while producing unstable event evidence. On DoTA Video Anomaly Recognition, Global Top-$K$ obtains competitive recognition and temporal grounding, but low selector--event alignment at $K=12$ (Frame AUC $51.5 \pm 10.8$). We propose dKFD, a phase-structured differentiable selector that reserves evidence capacity across pre-event, event, and post-event phases after full-sequence temporal encoding. Under matched-budget multi-seed evaluation, dKFD improves Frame AUC by $+30.97$ over a matched Global Top-$K$ selector at $K=12$ ($p<0.01$), while yielding modest but statistically significant recognition gains and comparable temporal grounding. Mechanism ablations show that phase supervision is load-bearing: removing it reduces Frame AUC to $41.1 \pm 12.0$ even when phase-partitioned budgets are retained. Downstream diagnostics on VRU-Accident show consistent gains over learned Global Top-$K$ across VLM families, while dense captioning reveals a boundary condition where uniform sampling remains competitive. These results support phase-structured allocation as a controlled fixed-budget approach for event-centric sparse evidence selection, not as a universal video summarization strategy.

\keywords{Sparse frame selection \and Video anomaly recognition \and Fixed-budget evidence allocation \and Temporal grounding \and Vision-language models}
\end{abstract}

\section{Introduction}
\label{sec:intro}
Long videos contain substantial redundancy despite advances in dense spatiotemporal video representation learning~\cite{tran2015c3d,wang2016tsn,carreira2017i3d,feichtenhofer2019slowfast}, but the evidence needed to understand a localized event is often sparse. Sparse frame selection is therefore a natural strategy for efficient video understanding, video summarization, and compact vision-language inputs~\cite{gong2014dpp,narasimhan2021clipit,wang2025seal,qian2024videostreaming,wu2024videollmmod,islam2025bimba}. However, sparse selection is often treated as generic compression: choose representative or salient frames and discard the rest. For temporally localized events, this view is incomplete. A frame is useful not only because it is visually distinctive, but because it plays a temporal role in explaining what happened.

This design is closely related to sparse summarization and adaptive sampling approaches that rank frames globally according to saliency, representativeness, or downstream utility~\cite{gong2014dpp,narasimhan2021clipit,tang2025aks}. This design is simple and strong, but it can be a poor inductive bias for localized events. Under recognition and grounding supervision, a global selector may concentrate on discriminative event peaks, preserving classification and coarse localization while under-selecting pre-event context and post-event consequences. We observe this failure mode on DoTA Video Anomaly Recognition~\cite{dota}: at $K=12$, a matched Global Top-$K$ selector reaches competitive recognition and grounding, yet its selector--event alignment is low and unstable across seeds.

We study this issue through the \emph{Phase-Structured Evidence Allocation Hypothesis}: under a fixed evidence budget, sparse event evidence should reserve capacity for pre-event context, event evidence, and post-event consequence, rather than allowing all frames to compete globally. This is a structural prior, not a claim that every event has clean phase boundaries or that fixed phase budgets are universally optimal.

We instantiate this hypothesis with dKFD, a phase-structured differentiable selector. dKFD first temporally encodes the full video sequence using a transformer-based temporal encoder~\cite{vaswani2017attention}, and then uses phase-specific selectors to allocate a fixed budget $K=K_{\mathrm{pre}}+K_{\mathrm{evt}}+K_{\mathrm{post}}$. The selected frames form an explicit evidence pack for recognition and inspection, while temporal grounding remains full-sequence and is conditioned on selector-derived cues.

Our main finding is that phase-structured allocation primarily improves evidence alignment, not boundary localization. Under matched backbone, budget, training protocol, and seeds, dKFD improves Frame AUC by $+30.97$ over Global Top-$K$ at $K=12$, while yielding modest recognition gains and comparable grounding. Across budgets $K \in \{8,12,16\}$, dKFD maintains stable Frame AUC around $81$--$82\%$, whereas Global Top-$K$ remains near chance with high variance. Mechanism ablations show that temporal encoding mainly supports grounding, while phase-aware selection and phase supervision drive evidence alignment.

We further evaluate the exported evidence packs on downstream VLM diagnostics. On VRU-Accident VQA, dKFD consistently improves over learned Global Top-$K$ across multiple VLM families, suggesting that phase-structured evidence can provide stronger compact inputs for downstream long-video reasoning systems~\cite{qian2024videostreaming,wu2024videollmmod,islam2025bimba}. However, dense captioning exposes a boundary: when the task rewards broad scene coverage rather than event-centric evidence, uniform sampling remains competitive. We therefore position dKFD as an event-centric fixed-budget evidence selector, not as a generic video summarizer.

Our contributions are:
\begin{enumerate}
    \item We identify a structural failure mode of globally competitive sparse selectors: they can preserve recognition and grounding while producing unstable event evidence.
    \item We propose and test the Phase-Structured Evidence Allocation Hypothesis under matched-budget multi-seed evaluation.
    \item We introduce dKFD, a phase-structured differentiable selector trained with recognition, grounding, and weak phase supervision.
    \item We provide controlled mechanisms and downstream diagnostics that clarify both the utility and boundary conditions of phase-structured evidence allocation.
\end{enumerate}


\section{Related Work}

\subsubsection{Sparse video selection and summarization.}
Video summarization and sparse frame selection aim to reduce visual redundancy by selecting representative, diverse, or salient frames from a longer sequence~\cite{gong2014dpp,narasimhan2021clipit,wang2025seal}. Recent learnable selectors such as Adaptive Keyframe Sampling further improve compact video representation by adapting selected frames to downstream tasks~\cite{tang2025aks}. These methods are valuable when the goal is broad coverage or efficient processing. Our setting differs because the selected frames must serve as event evidence: they should jointly support recognition, temporal grounding, and post-hoc inspection of a localized event.

\subsubsection{Temporal grounding and action localization.}
Temporal grounding and action localization methods predict event intervals, action boundaries, or query-relevant temporal segments~\cite{zhang2020tan,nan2021ivgdcl,li2021cpnet,moon2023qddetr,lin2023univtg,zhang2023tvprompt,dong2024relevancefeedback}. Strong temporal detectors such as ActionFormer and TriDet provide effective boundary prediction architectures~\cite{actionformer,tridet}. These models estimate when an event occurs, but a predicted interval is not the same as an inspectable fixed-budget evidence pack. dKFD is not intended to replace specialized localizers; instead, we study whether a compact set of explicit frames can be selected while preserving recognition, temporal grounding, and selector--event alignment.

\subsubsection{Efficient long-video and VLM inputs.}
Long-video vision-language models also face an information allocation problem under token constraints. Existing approaches reduce cost through streaming memory, adaptive computation, sparse visual processing, or selective compression~\cite{qian2024videostreaming,wu2024videollmmod,islam2025bimba}. In contrast, dKFD studies how a fixed number of explicit visual frames should be allocated when the event of interest is temporally localized. The resulting evidence packs contain frame indices, timestamps, phase labels, and selector scores, making the selected evidence directly inspectable.

\subsubsection{Positioning.}
Table~\ref{tab:positioning} summarizes the distinction between our setting and
nearby lines of work. The main difference is not only terminology, but also what
can be tested. Summarization evaluates coverage, localization evaluates
boundaries, and long-video VLM compression evaluates efficient reasoning under
token limits. In contrast, dKFD studies whether a fixed visual evidence budget
should be allocated across temporal roles for localized events, and whether phase
supervision is necessary for such allocation to become event-aligned.

\begin{table}[!htbp]
\centering
\small
\caption{
Positioning of dKFD relative to neighboring video-selection and long-video
understanding settings. dKFD studies supervised fixed-budget evidence allocation
for temporally localized events.
}
\label{tab:positioning}
\resizebox{\linewidth}{!}{
\begin{tabular}{p{0.20\linewidth} p{0.22\linewidth} p{0.25\linewidth} p{0.25\linewidth}}
\toprule
\textbf{Setting} &
\textbf{Selection objective} &
\textbf{What ``good'' selection means} &
\textbf{What this setting does not directly test} \\
\midrule
Video summarization &
Preserve broad video content &
Coverage, diversity, representativeness &
Whether supervised phase allocation produces event-aligned evidence under fixed $K$ \\
\midrule
Temporal grounding / action detection &
Predict event interval &
Accurate temporal boundaries &
Whether sparse, exportable evidence is preserved alongside localization \\
\midrule
Long-video VLM compression &
Reduce visual tokens for LLM reasoning &
Efficient answer generation under token limits &
Whether explicit visual-frame allocation under fixed $K$ benefits from phase structure \\
\midrule
\textbf{dKFD setting} &
\textbf{Allocate fixed-budget evidence} &
\textbf{Supervised pre/event/post evidence under fixed $K$} &
\textbf{This paper} \\
\bottomrule
\end{tabular}
}
\end{table}

\section{Problem Formulation}

We study sparse evidence selection for temporally localized events under a fixed frame budget. Given a video $X=\{x_t\}_{t=1}^{T}$, an event label $y$, and an annotated event interval $(t_s,t_e)$, the goal is to select a compact subset of frames $F \subset X$ such that $|F|\leq K$. The selected evidence should support three goals: recognizing the event category, localizing the event in time, and providing an inspectable set of frames that helps explain the event.

\subsubsection{Global Top-$K$ selection.}
A common sparse-selection strategy scores all frames and selects the $K$ highest-scoring ones:
\begin{equation}
I_{\mathrm{global}} = \mathrm{TopK}\left(\{s_t\}_{t=1}^{T}, K\right),
\qquad
F_{\mathrm{global}} = \{x_t : t \in I_{\mathrm{global}}\}.
\end{equation}
This is a valid and strong baseline, but it forces all frames to compete for one budget. For localized events, this can favor visually discriminative event peaks while discarding context before the event and consequences after it.

\subsubsection{Phase-structured evidence allocation.}
We instead test whether the fixed budget should be allocated across temporal roles:
\begin{equation}
K_{\mathrm{pre}} + K_{\mathrm{evt}} + K_{\mathrm{post}} = K,
\qquad
F = F_{\mathrm{pre}} \cup F_{\mathrm{evt}} \cup F_{\mathrm{post}}.
\end{equation}
The phase of a frame is relational: whether a frame is pre-event, event, or post-event depends on its position in the surrounding sequence, not only on its visual appearance. This motivates full-sequence temporal encoding before selection.

\subsubsection{Evaluation axes.}
Recognition accuracy alone is not sufficient for evaluating sparse evidence. A model can classify an event correctly while selecting frames that are poor evidence for inspection. We therefore evaluate each selector along three axes:
\begin{enumerate}
    \item \textbf{Recognition:} AVG Top-1 and Macro-F1.
    \item \textbf{Temporal grounding:} temporal IoU and absolute time error.
    \item \textbf{Selector-event alignment:} Frame AUC and hard-frame overlap with annotated event regions.
\end{enumerate}

\subsubsection{Matched-budget evaluation.}
Across primary comparisons, we hold fixed the visual backbone, total evidence budget, train/test split, optimization protocol, and random seeds. Significance is assessed using paired seed-level tests over the matched seed list. This isolates the structural comparison of interest: global frame competition versus phase-structured evidence allocation.

\section{Method}
\label{sec:method}

\begin{figure}[t]
    \centering
    \includegraphics[width=\textwidth]{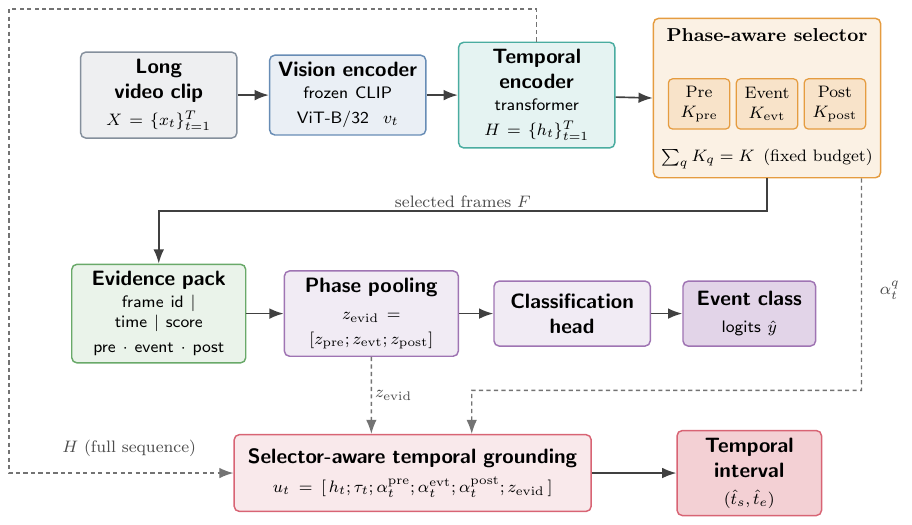}
    \caption{
    \textbf{dKFD overview.}
    The full video is temporally encoded before phase-aware selectors allocate a
    fixed evidence budget across pre-event, event, and post-event frames. The
    selected evidence pack is used for classification, while selector maps and
    pooled evidence condition temporal grounding over the full timeline.
    }
    \label{fig:method}
\end{figure}

dKFD instantiates phase-structured evidence allocation with a compact differentiable selector. The model first encodes the full video sequence, then allocates a fixed budget across pre-event, event, and post-event selectors. The selected frames form an explicit evidence pack for recognition and inspection, while temporal grounding remains full-sequence, conditioned on selector-derived cues.

\subsection{Full-Sequence Temporal Encoding}

Given video frames $X=\{x_t\}_{t=1}^{T}$, we extract frozen visual features $v_t \in \mathbb{R}^{d}$ and pass them through a temporal encoder:
\begin{equation}
h_{1:T} = \mathrm{TemporalEnc}(v_{1:T}),
\qquad
h_t \in \mathbb{R}^{d_h}.
\end{equation}
We use a transformer encoder with sinusoidal positional encodings. Encoding before selection allows each frame representation to carry temporal context, which is necessary because phase membership depends on the frame's relation to the event interval and surrounding sequence.

\subsection{Phase-Aware Differentiable Selection}

For each phase $q \in \{\mathrm{pre},\mathrm{evt},\mathrm{post}\}$, dKFD predicts a phase-specific selector score:
\begin{equation}
s_t^q = f_q(h_t).
\end{equation}
The total budget is partitioned as:
\begin{equation}
K_{\mathrm{pre}} + K_{\mathrm{evt}} + K_{\mathrm{post}} = K.
\end{equation}
Unless otherwise stated, we use $K_{\mathrm{pre}}/K_{\mathrm{evt}}/K_{\mathrm{post}} = 3/5/4$ for $K=12$.

Each phase selector produces a soft selector map and a hard Top-$K_q$ mask:
\begin{equation}
\alpha_t^q = \sigma(s_t^q/\tau),
\qquad
m_{1:T}^{q} = \mathrm{TopKMask}(s_{1:T}^{q}, K_q).
\end{equation}
The hard masks define the selected frames used in the forward pass and exported evidence pack, while gradients are routed through the soft maps using a straight-through estimator. The selected evidence for each phase is:
\begin{equation}
F_q = \{x_t : m_t^q = 1\},
\qquad
F = F_{\mathrm{pre}} \cup F_{\mathrm{evt}} \cup F_{\mathrm{post}}.
\end{equation}

\subsection{Evidence-Conditioned Recognition}

Recognition uses only the selected evidence. For each phase, selected frame features are pooled into a phase evidence vector:
\begin{equation}
z_q =
\frac{\sum_{t=1}^{T} m_t^q h_t}
{\sum_{t=1}^{T} m_t^q + \epsilon},
\qquad
q \in \{\mathrm{pre},\mathrm{evt},\mathrm{post}\}.
\end{equation}
The phase vectors are concatenated:
\begin{equation}
z_{\mathrm{evid}} = [z_{\mathrm{pre}}; z_{\mathrm{evt}}; z_{\mathrm{post}}],
\end{equation}
and the classifier predicts:
\begin{equation}
\hat{y} = g_{\mathrm{cls}}(z_{\mathrm{evid}}).
\end{equation}
This makes the exported evidence pack the recognition bottleneck.

\subsection{Selector-Aware Temporal Grounding}

Grounding predicts start and end distributions over the full timeline, but is conditioned on selector maps and the pooled evidence representation:
\begin{equation}
u_t =
[h_t; \tau_t; \alpha_t^{\mathrm{pre}}; \alpha_t^{\mathrm{evt}}; \alpha_t^{\mathrm{post}}; z_{\mathrm{evid}}],
\qquad
\tau_t = \frac{t}{T}.
\end{equation}
The grounding head predicts:
\begin{equation}
(\hat{p}_{\mathrm{start}}, \hat{p}_{\mathrm{end}}) = g_{\mathrm{loc}}(u_{1:T}).
\end{equation}
This design keeps recognition tied to sparse evidence while allowing localization to use full-sequence temporal context.

\subsection{Training Objective}

dKFD is trained with:
\begin{equation}
\mathcal{L}
=
\lambda_{\mathrm{cls}}\mathcal{L}_{\mathrm{cls}}
+
\lambda_{\mathrm{loc}}\mathcal{L}_{\mathrm{loc}}
+
\lambda_{\mathrm{budget}}\mathcal{L}_{\mathrm{budget}}
+
\lambda_{\mathrm{phase}}\mathcal{L}_{\mathrm{phase}}
+
\lambda_{\mathrm{div}}\mathcal{L}_{\mathrm{div}}
+
\lambda_{\mathrm{smooth}}\mathcal{L}_{\mathrm{smooth}}.
\end{equation}
Here, $\mathcal{L}_{\mathrm{cls}}$ supervises event recognition, $\mathcal{L}_{\mathrm{loc}}$ supervises temporal grounding, and $\mathcal{L}_{\mathrm{phase}}$ aligns phase selectors with weak pre-event, event, and post-event masks derived from the annotated interval. The remaining terms encourage budget adherence, selection diversity, and temporal smoothness. Full definitions and hyperparameters are provided in the supplementary material.

\subsection{Evidence Pack Export}

At inference time, dKFD exports:
\begin{equation}
E = \{(t, x_t, q, s_t^q) : m_t^q = 1,\ q \in \{\mathrm{pre},\mathrm{evt},\mathrm{post}\}\}.
\end{equation}
Each selected frame is associated with its frame index, timestamp, phase label, and selector score. This makes the selected evidence directly inspectable and enables matched-budget comparison with other selection policies.

\section{Experimental Setup}

\subsection{Datasets}

Our primary controlled benchmark is DoTA Video Anomaly Recognition~\cite{dota}, which provides event labels and temporal event intervals. This allows us to jointly evaluate recognition, temporal grounding, and selector-event alignment under a fixed frame budget. We follow the official train/validation/test split.

We use VRU-Accident~\cite{kim2025vruaccident} and DAD~\cite{chan2016anticipating} as diagnostic benchmarks. VRU-Accident evaluates whether exported evidence packs are useful as compact inputs to off-the-shelf VLMs. DAD evaluates whether phase-pretrained selectors transfer to accident anticipation. These datasets are not treated as equally controlled primary benchmarks. We report three metric groups: (1) \textbf{Recognition:} official AVG Top-1 and Macro-F1; (2) \textbf{Temporal grounding:} temporal IoU and absolute event-center time error; and (3) \textbf{Selector--event alignment:} Frame AUC, computed on held-out clips using annotated event intervals.

\subsection{Baselines and Controls}

The primary baseline is a matched Global Top-$K$ selector. Both dKFD and Global Top-$K$ use the same frozen visual features, temporal encoder, total budget, training protocol, and seed list. The key difference is the selection structure: dKFD partitions the budget across phases, while Global Top-$K$ lets all frames compete globally.

We also compare against pooling controls, temporal-localization baselines adapted to the same frozen features, and heuristic structured selection. These contextual baselines help distinguish phase-aligned evidence selection from generic temporal coverage or boundary prediction.

\section{Controlled DoTA Results}

\subsection{Global Top-$K$ Preserves Tasks but Not Stable Evidence}

Table~\ref{tab:main_dota} reports the primary five-seed comparison at $K=12$. Global Top-$K$ achieves competitive recognition and grounding, with $30.46 \pm 0.42$ AVG Top-1 and $53.22 \pm 0.62$ tIoU. However, its selector--event alignment is weak and unstable: Frame AUC is $51.48 \pm 10.81$. This shows that a sparse selector can support classification and grounding while still producing unreliable event evidence. Figure~\ref{fig:evidence}(a) makes this spread visible: the Global Top-$K$ band varies by more than $30$ AUC points across the matched seed list and straddles chance, whereas dKFD stays tightly clustered --- the instability is structural rather than incidental.

\subsection{Phase-Structured Allocation Improves Evidence Alignment}

Under the same frozen features, total budget, protocol, and seed list, dKFD substantially improves the main failure mode. Frame AUC increases from $51.48 \pm 10.81$ to $82.45 \pm 0.65$, a gain of $+30.97$ points with paired seed-level significance. Recognition also improves, but the gains are modest: $+1.64$ AVG Top-1 and $+2.23$ Macro-F1. Temporal grounding remains comparable. The main effect is therefore not sharper boundary localization, but more reliable event-aligned evidence under the same fixed budget.

\begin{table}[t]
\centering
\caption{Main controlled DoTA comparison at $K=12$ over five seeds. Asterisks indicate paired seed-level significance against Global Top-$K$: $^{**}p<0.01$, $^{***}p<0.001$.}
\label{tab:main_dota}
\resizebox{\linewidth}{!}{
\begin{tabular}{lccccc}
\toprule
Method & AVG Top-1 $\uparrow$ & Macro-F1 $\uparrow$ & tIoU $\uparrow$ & Time Err. $\downarrow$ & Frame AUC $\uparrow$ \\
\midrule
Global Top-$K$ & $30.46 \pm 0.42$ & $28.25 \pm 0.85$ & $53.22 \pm 0.62$ & $9.41 \pm 0.21$ & $51.48 \pm 10.81$ \\
dKFD & $32.10^{**} \pm 0.75$ & $30.48^{**} \pm 0.52$ & $53.69 \pm 0.45$ & $9.18 \pm 0.21$ & $82.45^{**} \pm 0.65$ \\
\midrule
$\Delta$ & $+1.64$ & $+2.23$ & $+0.47$ & $-0.23$ & $+30.97$ \\
\bottomrule
\end{tabular}
}
\end{table}

\subsection{Budget Scaling}

\begin{figure}[t]
\centering
\includegraphics[width=\textwidth]{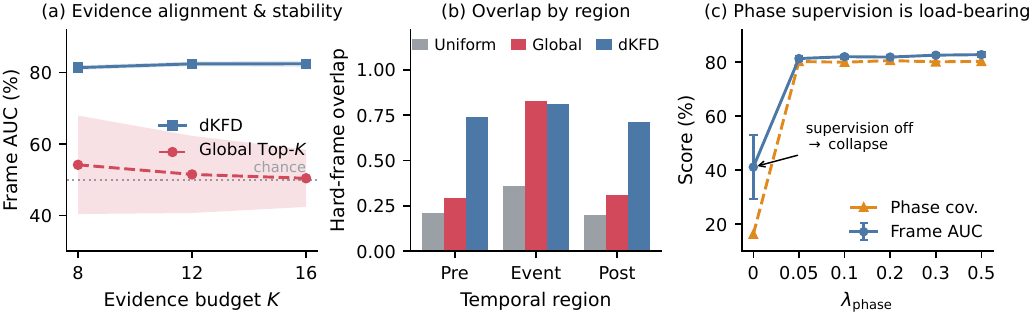}
\caption{\textbf{Evidence-quality summary on DoTA.}
(a) Frame AUC versus evidence budget $K$: dKFD stays tightly clustered near
$82\%$ across budgets, whereas Global Top-$K$ shows a wide per-seed spread
(shaded $\pm$std) straddling chance --- capturing both budget-robustness and
seed instability in a single view. (b) Hard-frame overlap by temporal region at
$K=12$: dKFD's gain over Global Top-$K$ comes from the pre- and post-event
regions, not from over-selecting the event window. (c) Phase-supervision sweep:
removing phase supervision ($\lambda_{\mathrm{phase}}=0$) collapses both Frame
AUC and phase coverage, while moderate supervision recovers stable alignment ---
phase supervision is load-bearing.}
\label{fig:evidence}
\end{figure}

Table~\ref{tab:budget_scaling} repeats the matched comparison for $K \in \{8,12,16\}$; Figure~\ref{fig:evidence}(a) visualizes the Frame AUC trend across budgets. Across all budgets, dKFD maintains high and stable Frame AUC in the $81$--$82\%$ range, while Global Top-$K$ remains near $50$--$54\%$ with substantially larger variance. Recognition gains are positive across budgets but reach significance only at the headline $K=12$ setting. We therefore avoid claiming monotonic scaling; the supported conclusion is that phase-structured allocation produces more reliable event evidence than global competition under matched fixed budgets.

\begin{table}[t]
\centering
\caption{Budget scaling under matched frozen features and five-seed evaluation. Asterisks indicate paired significance against matched Global Top-$K$ at the same $K$: $^{*}p<0.05$, $^{**}p<0.01$, $^{***}p<0.001$.}
\label{tab:budget_scaling}
\begin{tabular}{lcccc}
\toprule
Setting & AVG Top-1 $\uparrow$ & Macro-F1 $\uparrow$ & tIoU $\uparrow$ & Frame AUC $\uparrow$ \\
\midrule
Global, $K=8$ & $30.34 \pm 1.47$ & $28.75 \pm 1.42$ & $52.96 \pm 0.80$ & $54.17 \pm 13.79$ \\
dKFD, $K=8$ & $31.84 \pm 1.88$ & $30.11 \pm 1.40$ & $53.24 \pm 0.38$ & $81.39^{*} \pm 0.62$ \\
Global, $K=12$ & $30.46 \pm 0.42$ & $28.25 \pm 0.85$ & $53.22 \pm 0.62$ & $51.48 \pm 10.81$ \\
dKFD, $K=12$ & $32.10^{**} \pm 0.75$ & $30.48^{**} \pm 0.52$ & $53.69 \pm 0.45$ & $82.45^{**} \pm 0.65$ \\
Global, $K=16$ & $30.33 \pm 1.35$ & $27.99 \pm 0.57$ & $53.54 \pm 0.65$ & $50.39 \pm 8.02$ \\
dKFD, $K=16$ & $31.51 \pm 1.40$ & $29.72 \pm 1.30$ & $53.75 \pm 0.52$ & $82.47^{***} \pm 0.85$ \\
\bottomrule
\end{tabular}
\end{table}

\subsection{DAD Transfer Diagnostic}

We evaluate phase- and global-pretrained selectors on DAD under fine-tuning and scratch training. We treat DAD as an external transfer diagnostic rather than a primary anticipation benchmark. Under fine-tuning, phase-pretrained selectors improve over global-pretrained selectors on AUC, AP, and FAR. Under scratch training, phase and global architectures are effectively tied. This suggests that phase structure alone is not sufficient; the phase-supervised pretraining signal is what makes the selector behave as event-aligned evidence. Full DAD results are provided in the supplementary material.

\section{Mechanism and Sensitivity Analysis}

\subsection{Component Roles Are Separable}

Table~\ref{tab:mechanism} shows that different components of dKFD affect different evaluation axes. Removing the temporal encoder preserves recognition but severely hurts grounding, reducing tIoU from $53.69$ to $39.86$. In contrast, removing phase heads preserves grounding but collapses Frame AUC from $82.16$ to $48.66$. This separation supports the main mechanism: temporal context is important for credible interval prediction, while phase-aware selection is central to event-aligned evidence.

\begin{table}[t]
\centering
\caption{Mechanism ablations on DoTA over five seeds. Temporal encoding mainly supports grounding, while phase-aware heads mainly support selector--event alignment.}
\label{tab:mechanism}
\begin{tabular}{p{2.3cm}p{1.9cm}p{1.7cm}p{1.7cm}p{1.7cm}p{1.9cm}}
\toprule
Variant & AVG Top-1 $\uparrow$ & Macro-F1 $\uparrow$ & tIoU $\uparrow$ & Time Err. $\downarrow$ & Frame AUC $\uparrow$ \\
\midrule
Full dKFD & $31.94 \pm 0.23$ & $30.28 \pm 0.44$ & $53.69 \pm 0.45$ & $9.11 \pm 0.18$ & $82.16 \pm 0.69$ \\
w/o temporal encoder & $32.32 \pm 1.06$ & $30.54 \pm 0.89$ & $39.86 \pm 0.92$ & $14.32 \pm 0.24$ & $66.00 \pm 0.60$ \\
w/o phase heads & $30.17 \pm 1.34$ & $28.28 \pm 1.42$ & $53.63 \pm 0.42$ & $9.35 \pm 0.07$ & $48.66 \pm 9.71$ \\
w/o selector-aware grounding & $30.80 \pm 1.40$ & $29.24 \pm 1.22$ & $52.98 \pm 0.56$ & $9.35 \pm 0.18$ & $82.70 \pm 0.57$ \\
w/o evidence summary & $31.17 \pm 1.09$ & $29.53 \pm 0.87$ & $53.30 \pm 0.96$ & $9.24 \pm 0.16$ & $82.41 \pm 0.14$ \\
w/o selector maps & $32.17 \pm 1.31$ & $30.21 \pm 0.95$ & $53.55 \pm 0.40$ & $9.15 \pm 0.07$ & $81.93 \pm 0.46$ \\
\bottomrule
\end{tabular}
\end{table}

\subsection{Phase Supervision Is Load-Bearing}

Table~\ref{tab:phase_sweep} tests whether the gain comes from phase partitioning alone or from phase supervision. When $\lambda_{\mathrm{phase}}=0$, Frame AUC drops to $41.12 \pm 11.99$ even though phase-partitioned budgets and joint recognition--grounding training remain. Phase coverage also collapses from roughly $80\%$ to $16.3\%$. With moderate phase supervision, Frame AUC recovers above $81\%$ across the sweep. Thus, the architecture creates capacity to reserve evidence across temporal roles, but phase supervision is what aligns that capacity with event structure.

\begin{table}[t]
\centering
\caption{Sensitivity to phase supervision over three seeds. Without phase supervision, Frame AUC and phase coverage drop sharply despite retaining phase-partitioned budgets.}
\label{tab:phase_sweep}
\begin{tabular}{lccccc}
\toprule
$\lambda_{\mathrm{phase}}$ & AVG Top-1 $\uparrow$ & Macro-F1 $\uparrow$ & tIoU $\uparrow$ & Frame AUC $\uparrow$ & Phase cov. $\uparrow$ \\
\midrule
$0$ & $30.78 \pm 2.25$ & $28.98 \pm 1.34$ & $54.40 \pm 1.01$ & $41.12 \pm 11.99$ & $16.3 \pm 15.3$ \\
$0.05$ & $30.17 \pm 0.95$ & $28.71 \pm 0.29$ & $53.60 \pm 0.64$ & $81.34 \pm 0.94$ & $80.2 \pm 0.9$ \\
$0.1$ & $30.34 \pm 0.64$ & $28.79 \pm 0.40$ & $53.34 \pm 1.27$ & $82.02 \pm 0.66$ & $80.0 \pm 1.4$ \\
$0.2$ & $31.95 \pm 0.56$ & $30.26 \pm 0.44$ & $53.42 \pm 0.53$ & $81.88 \pm 0.67$ & $80.6 \pm 1.0$ \\
$0.3$ & $30.97 \pm 1.00$ & $29.42 \pm 1.29$ & $53.65 \pm 0.37$ & $82.59 \pm 0.67$ & $80.1 \pm 1.4$ \\
$0.5$ & $30.88 \pm 1.01$ & $29.41 \pm 1.17$ & $54.45 \pm 0.23$ & $82.78 \pm 0.86$ & $80.3 \pm 0.6$ \\
\bottomrule
\end{tabular}
\end{table}

\subsection{Selector Diagnostics}

Additional diagnostics support the same interpretation. Classification-only training reaches non-trivial recognition but fails on grounding and Frame AUC, confirming that classification accuracy is not a reliable proxy for evidence quality. Full dKFD selects $11.97$ frames on average for nominal $K=12$, while Global Top-$K$ selects $12.00$, so the Frame AUC gain is not explained by selecting more frames. Decomposing hard-frame overlap by temporal region (Figure~\ref{fig:evidence}(b)) shows that dKFD's improvement over Global Top-$K$ comes specifically from the pre- and post-event regions, while event-region overlap is comparable across methods --- a structurally non-circular complement to Frame AUC. Additional gate-level AUC diagnostics are reported in the supplementary material.

\subsection{Qualitative Evidence Allocation}

\begin{figure}[t]
\centering
\includegraphics[width=\textwidth, trim=0 243 0 0, clip]{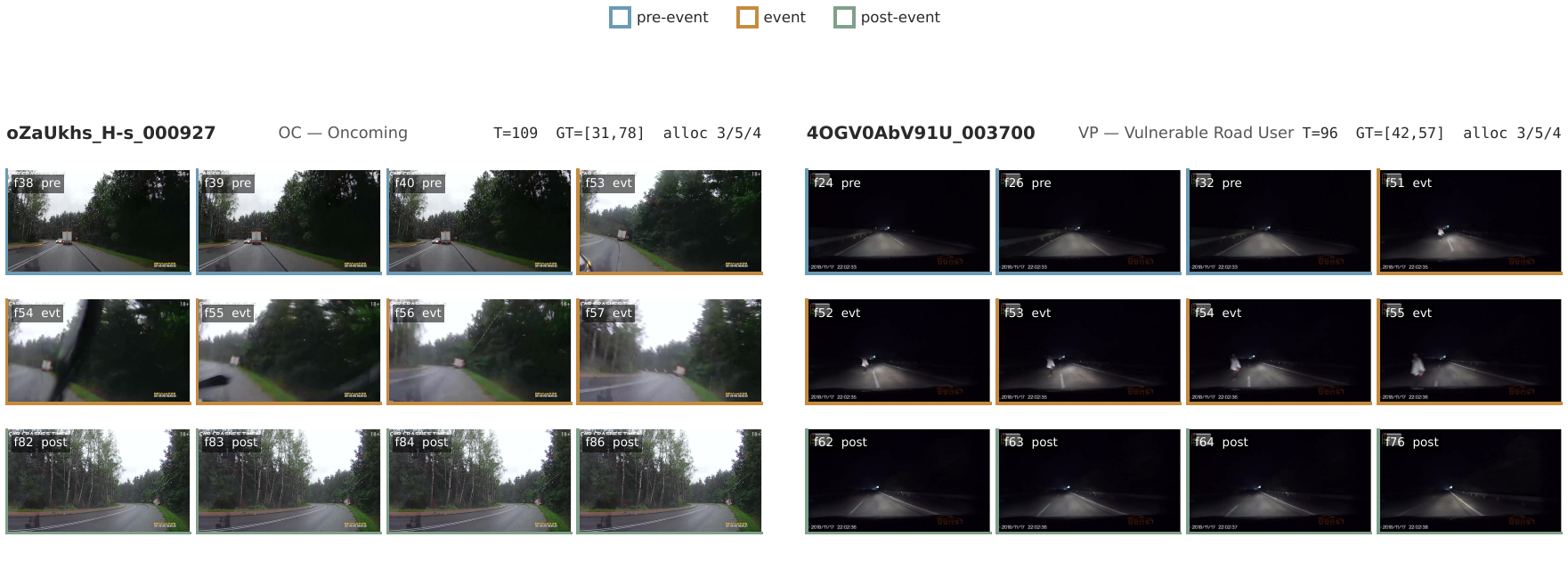}
\caption{\textbf{Qualitative selector comparison at $K=12$ on two DoTA test
clips} (left: OC, oncoming; right: VP, vulnerable road user). The cream band
marks the annotated event interval; the three lanes per clip show Uniform
sampling, Global Top-$K$, and dKFD. Uniform covers the timeline but is not
task-supervised; Global Top-$K$ concentrates frames near the event core; dKFD
allocates the same budget across pre-event, event, and post-event phases,
covering causal setup, interaction, and aftermath.}
\label{fig:qualitative}
\end{figure}

We visualize selected evidence for Uniform sampling, Global Top-$K$, and dKFD at the same budget in Figure~\ref{fig:qualitative}; we encourage readers to inspect this figure as it provides the most direct evidence of the method's interpretability. Uniform sampling covers the timeline but is not task-supervised. Global Top-$K$ often concentrates selected frames near the event core. dKFD instead reserves capacity for pre-event setup, event evidence, and post-event aftermath. These qualitative examples make the selector--event alignment improvement visible rather than only numerical. The supplementary material further visualizes the soft phase-selector activations on representative clips: each phase head fires in its expected temporal region, providing direct evidence that the phase decomposition is a learned property of the selector rather than imposed at inference time.

\section{Downstream Utility and Boundary Conditions}

\subsection{VRU-Accident VQA}

We evaluate dKFD as a zero-shot frame-selection policy for VRU-Accident VQA using 6,000 questions. The selected frames are passed to off-the-shelf VLMs without retraining. Table~\ref{tab:vru_qwen} shows results for Qwen2.5-VL-3B~\cite{bai2025qwen25vl} at $K=12$. dKFD improves over learned Global Top-$K$ on the overall average and on event-grounded categories such as Accident Type, Accident Cause, and Prevention Method. However, the comparison against uniform sampling is mixed: dKFD is slightly better for Qwen2.5-VL-3B, comparable for LLaVA-OneVision~\cite{li2024llavaonevision}, and worse for InternVL3-2B~\cite{zhu2025internvl3}. We therefore make a bounded claim: dKFD provides structured event-centric evidence that is consistently better than learned Global Top-$K$, but it does not universally dominate coverage-based sampling. A per-VLM comparison across all three VLM families is provided in the supplementary material.

\begin{table}[t]
\centering
\caption{VRU-Accident VQA on Qwen2.5-VL-3B at $K=12$ using 6,000 questions and the official evaluator. Extended multi-VLM results are provided in the supplementary material.}
\label{tab:vru_qwen}
\begin{tabular}{lccccccc}
\toprule
Input & WL & TE & RC & AT & AC & PM & AVG $\uparrow$ \\
\midrule
Full video & $63.80$ & $72.10$ & $39.40$ & $47.30$ & $25.60$ & $44.70$ & $48.82$ \\
Global Top-$K$ & $66.90$ & $81.40$ & $38.20$ & $51.20$ & $26.80$ & $47.00$ & $51.92$ \\
Uniform & $69.60$ & $82.90$ & $36.30$ & $52.30$ & $28.60$ & $46.90$ & $52.77$ \\
dKFD & $67.90$ & $82.60$ & $38.40$ & $53.00$ & $28.20$ & $47.30$ & $52.90$ \\
\bottomrule
\end{tabular}
\end{table}

\subsection{Dense Captioning}

\begin{figure}[t]
\centering
\includegraphics[width=\textwidth]{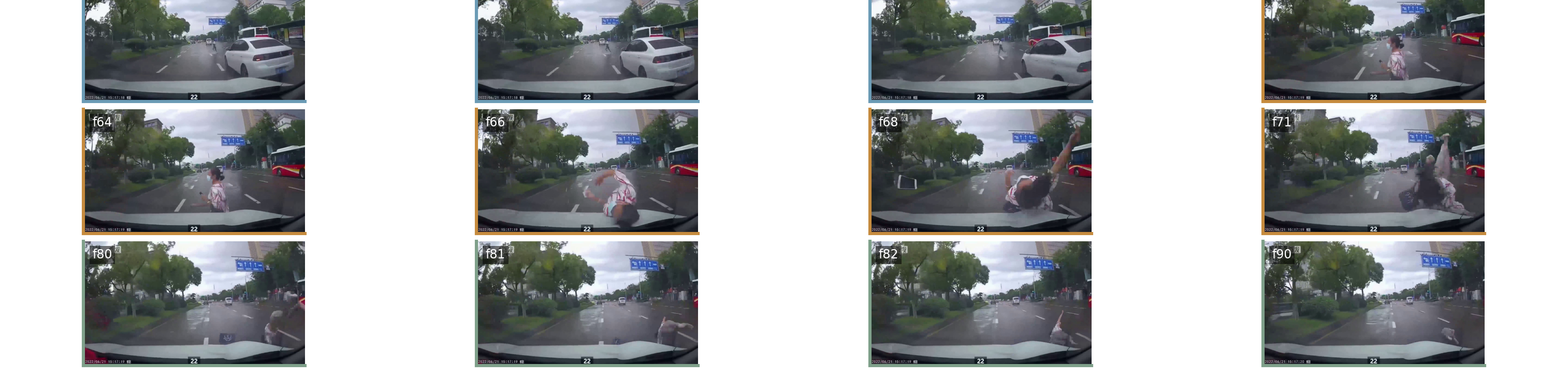}
\caption{\textbf{Dense-captioning use case on a VRU-Accident clip}
(``pedestrian runs out from behind an overtaken vehicle''). The $K{=}12$ dKFD
evidence pack shown above (frame borders mark pre/event/post phases) captures the
pedestrian's emergence, the crossing, the impact, and the aftermath; feeding
\emph{only} these frames to Qwen2.5-VL produces a coherent collision narrative.
\emph{Predicted (dKFD\,$\to$\,VLM):} ``The video captures a collision between a
car and a pedestrian on a city street\dots\ a white car is seen approaching\dots\
The pedestrian, dressed in a white shirt and dark pants, is crossing the street
in front of the car\dots\ As the car approaches, it makes a right turn onto the
next lane\dots\ the pedestrian's path intersects with the car's path, leading to
a collision.'' \emph{Ground truth (expert, abridged):} ``On an overcast
afternoon\dots\ a female pedestrian\dots\ suddenly runs from the right sidewalk
into the roadway at a diagonal angle\dots\ The driver's view of the pedestrian is
obstructed by the white vehicle\dots\ the driver fails to brake in time,
resulting in a collision\dots\ The vehicle continues traveling straight without
any apparent evasive action\dots''}
\label{fig:dense-example}
\end{figure}

Dense captioning exposes the boundary of phase-structured allocation. Unlike event-grounded VQA, dense captioning rewards broad whole-video description. In this setting, compact $K=12$ inputs improve over full-video input for Qwen2.5-VL, suggesting that removing redundant frames can help. However, uniform sampling and dKFD are effectively tied across captioning metrics. This is consistent with the task: dense captioning benefits from temporal coverage, while dKFD intentionally prioritizes event-centric evidence. Even so, the dKFD evidence pack remains a useful compact input: Figure~\ref{fig:dense-example} shows that, from only $K=12$ phase-structured frames, an off-the-shelf VLM recovers the core collision narrative of a VRU-Accident clip.

\section{Limitations}

Our study has several limitations. DoTA VAR is the only primary controlled benchmark. VRU-Accident and DAD are used as downstream and transfer diagnostics rather than equally controlled evaluations of the same recognition--grounding--evidence selection problem. Both DoTA and DAD are driving-accident datasets, so the hypothesis should be tested in other localized-event domains. dKFD uses a simple fixed phase budget. This prior will not fit every event. Some events have long buildup, abrupt onset, weak aftermath, or ambiguous phase boundaries. Adaptive or class-conditional phase budgets are a natural next step.

Additionally, phase supervision is load-bearing. When $\lambda_{\mathrm{phase}}=0$, Frame AUC drops sharply even though phase-partitioned budgets remain. As implemented here, dKFD requires event-interval annotations to construct weak phase masks. Semi-supervised, self-supervised, or pseudo-labeled phase supervision remains future work. Also, Frame AUC is only a proxy for evidence quality. We supplement it with hard-frame overlap, phase coverage, and qualitative grids, but we do not conduct a human evaluation of whether selected frames are more useful for inspection or explanation. Finally, dKFD is not a general video summarizer or a replacement for specialized temporal localizers. It is designed for fixed-budget event evidence selection, where preserving pre-event context, event evidence, and post-event consequence is useful.

\section{Conclusion}

We studied sparse visual evidence allocation for temporally localized events under a fixed frame budget. Our results show that globally competitive Top-$K$ selection can preserve recognition and grounding while producing unstable event evidence. We tested phase-structured allocation as a controlled alternative: instead of letting all frames compete for one budget, dKFD reserves capacity for pre-event, event, and post-event evidence. Under matched-budget multi-seed evaluation on DoTA Video Anomaly Recognition, dKFD improves Frame AUC by $+30.97$ over a matched Global Top-$K$ selector at $K=12$, while yielding modest recognition gains and comparable temporal grounding. Mechanism ablations show that temporal encoding, phase-aware selection, and selector-aware supervision serve different roles. Most importantly, phase supervision is load-bearing: without it, Frame AUC collapses even when phase-partitioned budgets and joint training are retained. Downstream diagnostics clarify the scope of the method. On VRU-Accident VQA, dKFD improves over learned Global Top-$K$ across VLM families. On dense captioning, uniform sampling remains competitive because the task rewards broad scene coverage rather than localized causal evidence. We therefore view dKFD not as a universal video summarizer, but as a controlled approach to fixed-budget evidence selection for localized events.

%
%
\bibliographystyle{splncs04}
\bibliography{main}

\end{document}